\documentclass[10pt,twocolumn,letterpaper]{article}

\usepackage{cvpr}
\usepackage{times}
\usepackage{epsfig}
\usepackage{graphicx}
\usepackage{amsmath}
\usepackage{amssymb}
\usepackage{bm}
\usepackage{multirow}
\usepackage{diagbox}

\usepackage[pagebackref=true,breaklinks=true,letterpaper=true,colorlinks,bookmarks=false]{hyperref}

\cvprfinalcopy 

\def\cvprPaperID{9112} 

\ifcvprfinal\fi

\begin{document}

\title{STAViS: Spatio-Temporal AudioVisual Saliency Network}

\author{Antigoni Tsiami, Petros Koutras and Petros Maragos \\
School of E.C.E., National Technical University of Athens, Greece\\
{\tt\small \{antsiami, pkoutras, maragos\}@cs.ntua.gr}
}


\maketitle

\begin{abstract}
We introduce STAViS\footnote{This research has been cofinanced by the European Union and Greek national funds through the Operational Program Competitiveness, Entrepreneurship and Innovation, under the call RESEARCH CREATE INNOVATE (project code:T1EDK- 01248 / MIS: 5030856).}, a spatio-temporal audiovisual saliency network that combines spatio-temporal visual and auditory information in order to efficiently address the problem of saliency estimation in videos. Our approach employs a single network that combines visual saliency and auditory features and learns to appropriately localize sound sources and to fuse the two saliencies in order to obtain a final saliency map. The network has been designed, trained end-to-end, and evaluated on six different databases that contain audiovisual eye-tracking data of a large variety of videos. We compare our method against 8 different state-of-the-art visual saliency models. Evaluation results across databases indicate that our STAViS model outperforms our visual only variant as well as the other state-of-the-art models in the majority of cases. Also, the consistently good performance it achieves for all databases indicates that it is appropriate for estimating saliency ``in-the-wild". The code is available at \href{https://github.com/atsiami/STAViS}{https://github.com/atsiami/STAViS}.
\end{abstract}

\vspace{-0.30cm}
\section{Introduction}

%
%
Audio is present in the majority of scenes where vision is also present.
Nature as well as everyday life offer unlimited examples of multi-sensory and cross-modal events, a large amount of which concern auditory and visual senses. Audiovisual integration is a well studied phenomenon in neuroscience with interesting findings that have motivated research in the area~\cite{MeredStein83,MeredStein86,vatakisSpence07}. Many of them provide sufficient evidence that human attention and perception is influenced by audiovisual stimuli in a different way than auditory or visual stimuli in isolation. For example, the well-known McGurk effect~\cite{mcgurk} proved that attention can be heavily affected by incongruent audiovisual stimuli, leading to illusionary perceptions. On the other hand, congruent ones, like in the pip \& pop effect~\cite{Van+08}, can lead to enhanced attention.
Modeling human attention is important in a variety of applications, like video summarization~\cite{susi,Eva+13}, robotics~\cite{dang2018visual,kovacs2019saliency,yuan2018rgb}, and virtual reality~\cite{sitzmann2018saliency}.

\begin{figure}[t]
\begin{center}
\includegraphics[width=\linewidth]{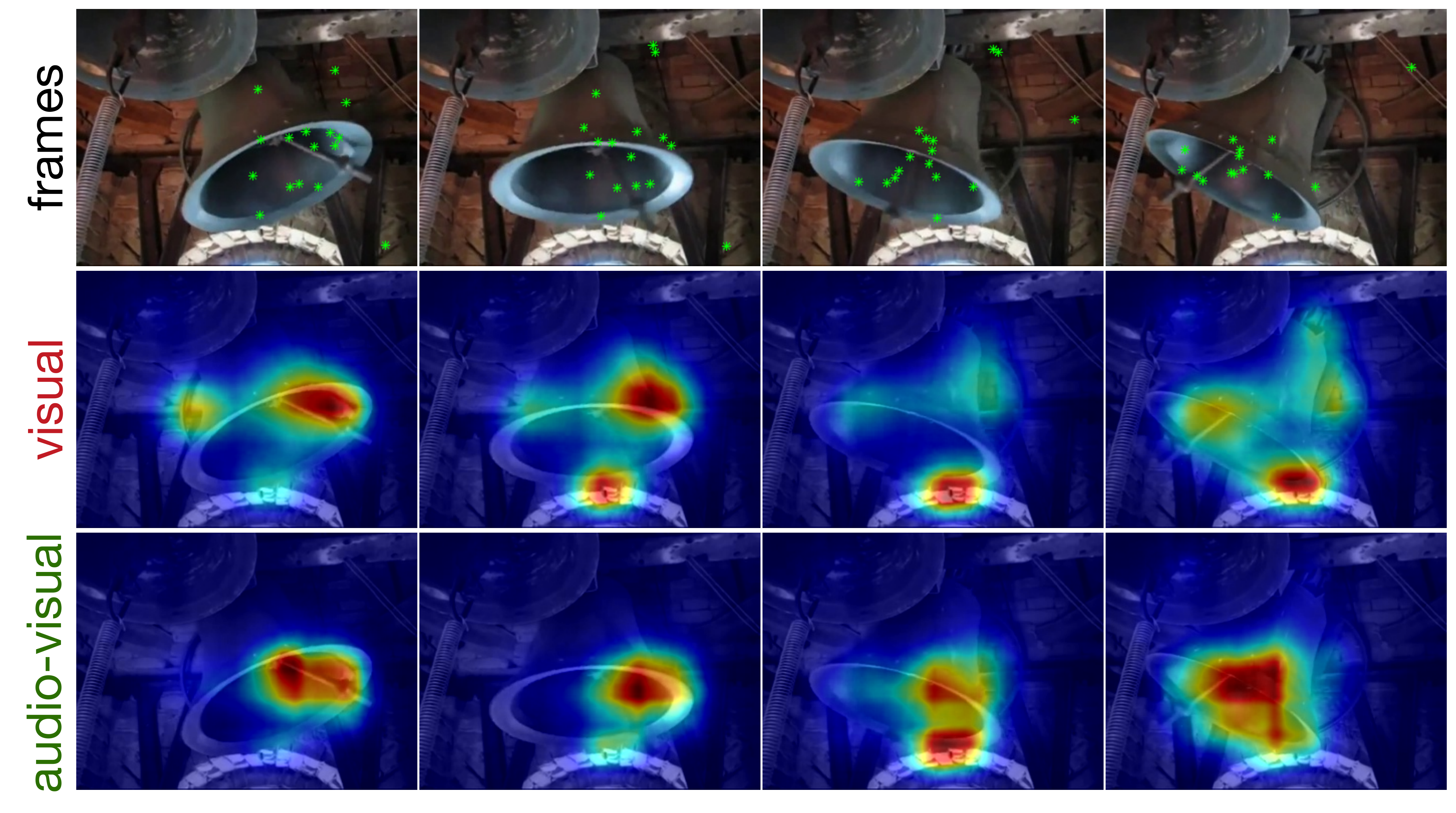}
\end{center}
\vspace{-0.4cm}
   \caption{Example frames with their eye-tracking data of a bell tolling. The second row depicts the saliency maps produced by our visual-only saliency network, while the third row is the output of our proposed STAViS network, which succeeds in better capturing human attention.}
\label{fig:front_fig}
\vspace{-0.4cm}
\end{figure}

Despite the previously mentioned evidence of a strong audiovisual interplay, audio modality has been rather neglected when modeling human attention, with most computer vision research focusing only on visual information. We aim to address the problem of saliency estimation in videos ``in-the-wild", namely how bottom-up human attention is modeled, by integrating auditory information into visual saliency, without prior knowledge of the video content. Fig.~\ref{fig:front_fig} highlights and motivates the problem we aim to address: tolling bells is an audiovisual event. The depicted frames, with their eye-tracking data superimposed, exhibit an example of human attention in an audiovisual scene. This attention can be much better captured by an audiovisual attention model than by a visual-only one, as the corresponding saliency maps indicate.

In parallel, Convolutional  Neural Networks (CNNs) have infiltrated computer vision and have enhanced dramatically the performance in the  majority  of  spatial tasks in computer vision, such as object detection or semantic segmentation \cite{SiZi15,he2016deep,maskRCNN}. The huge amounts of video data that are becoming more and more available through online sources, enable and at the same time require continuously better performance in tasks like activity recognition, video saliency, scene analysis or video summarization, imposing the need to exploit not only spatial information, but also temporal \cite{carreira2017quo,ray2018scenes,susi}. Similar advances have also been achieved in audio processing areas, such as acoustic event detection~\cite{primus2019exploiting}, speech recognition~\cite{graves2013speech, chiu2018state}, sound localization~\cite{tian2018audio}, by using deep learning techniques. 

As an additional free of charge source of information, auditory stream has been incorporated in many video-related applications during the last three years~\cite{arandjelovic2017look,zhao2018sound}. Audio contains rich information that when integrated with spatio-temporal visual information, it facilitates learning and improves performance in many traditionally visual-only tasks. 
Most importantly, audio comes for free, as it is already included in the majority of videos and synchronized with the visual stream. We are particularly interested in audiovisual attention modeling, through audiovisual saliency, where the main goal is to predict human fixations in a video.


We propose STAViS,  a novel spatio-temporal audiovisual saliency network that combines spatio-temporal visual with auditory information to efficiently address the problem of saliency estimation in videos. We mainly aspire to make the next logical step in video saliency prediction by introducing a new audiovisual framework rather than propose just a better visual model. To our best knowledge, our proposed system is the first deep learning saliency approach that employs both video and audio and addresses the fixation prediction problem, except for concurrent work~\cite{tavakoli2019dave} where two independent networks for the two modalities are employed and their outputs are simply concatenated as a late fusion scheme.
Our approach employs a single multimodal network that combines visual and auditory information at multiple stages and learns to appropriately fuse the two modalitites, in order to obtain a final saliency map. We propose ways to perform sound source localization in the video and subsequently fuse spatio-temporal auditory saliency with spatio-temporal visual saliency, in order to obtain the final audiovisual saliency map.

For visual saliency our approach is extending a state-of-the-art spatio-temporal saliency network, called SUSiNet~\cite{susi}, part of a multi-task network that jointly addresses 3 different problems: saliency estimation, activity recognition and video summarization. Regarding audio, we obtain audio features using SoundNet~\cite{aytar2016soundnet}, a state-of-the-art CNN for acoustic event classification. 
For sound localization, we aim to retrieve the potential correspondences between auditory and visual streams, in the sense of cross-modal semantic concepts. We explore 3 different ways of locating sounds in a video, obtaining an auditory saliency map. Subsequently, we investigate fusion schemes to integrate spatio-temporal and auditory saliencies and obtain a final saliency map. We explore three alternatives as well.

The network has been designed, trained end-to-end and evaluated on 6 different databases that contain audiovisual eye-tracking data. These databases contain a large variety of videos, ranging from home-made videos to documentaries, and from short, simple videos to Hollywood movies. Our method is compared to 8 different state-of-the-art visual saliency models in all 6 databases. Evaluation results across databases indicate that STAViS model outperforms both the visual variant network and the other methods in the majority of cases. Also, its consistently good performance for all databases, indicates that it is suitable for estimating saliency in videos ``in-the-wild".   

\begin{figure*}[t]
\begin{center}
\includegraphics[width=\textwidth]{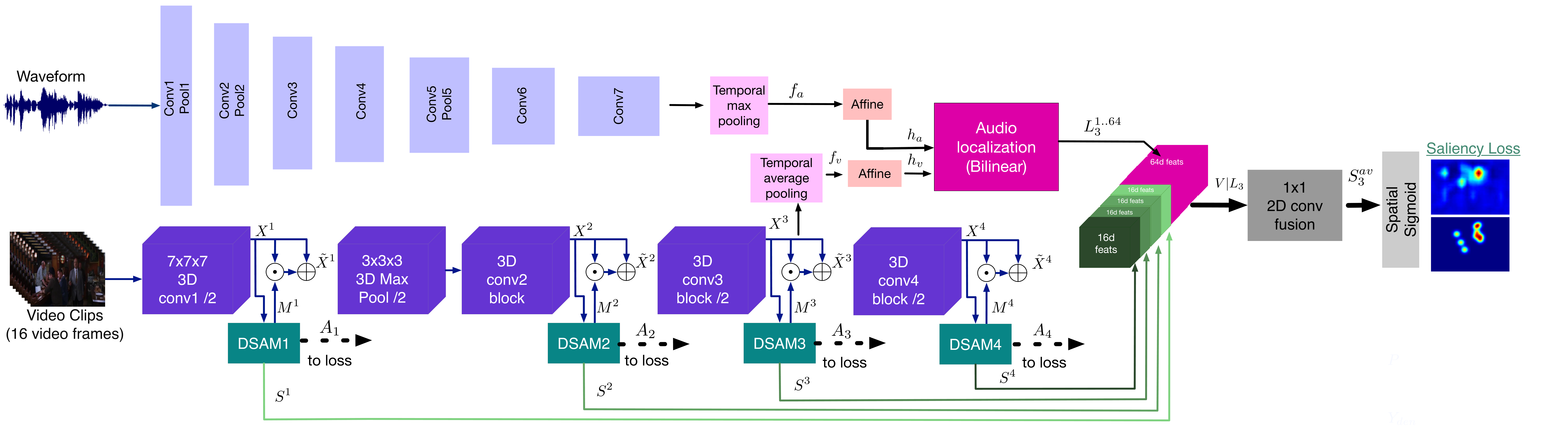}
\end{center}
\vspace{-0.2cm}
   \caption{STAViS architecture: the spatio-temporal audiovisual network is based on the ResNet architecture and has one spatio-temporal visual path, one auditory path and their fusion.}
\label{fig:susi}
\end{figure*}

\section{Related Work}

There is a very broad literature regarding visual saliency, both for classical approaches and for deep learning ones. On the other hand, audiovisual approaches in saliency are limited. Below, we describe the most notable works in visual saliency with deep methods, audiovisual saliency, and other recent audiovisual approaches in similar problems.

\noindent\textbf{Visual Saliency:} 
The early CNN-based approaches for saliency were based on the adaptation of pretrained CNN models for visual recognition tasks \cite{Kummerer2014b,vig2014large}. Later, in \cite{Pan_2016_CVPR} both shallow and deep CNN were trained end-to-end for saliency prediction while \cite{huang2015salicon, jetley2016end} trained the networks by optimizing common saliency evaluation metrics. In \cite{Pan_2017_SalGAN} the authors employed end-to-end Generative Adversarial Networks (GAN), while \cite{wang2018deep} has utilized multi-level saliency information from different layers through skip connections. Long Short-term Memory (LSTM) networks have also been used for tracking visual saliency both in static images \cite{cornia2018predicting} and video stimuli \cite{wang2018revisiting,wang2019revisiting,Linardos2019}. In order to improve saliency estimation in videos, many approaches employ multi-stream networks, such as RGB/Optical Flow (OF) \cite{bak2017spatio,lai2019video}, RGB/OF/Depth \cite{leifman2017learning}, or multiple subnets such as objectness/motion \cite{jiang2018deepvs} or saliency/gaze \cite{Gorji_2018_CVPR} pathways.
Another recent approach~\cite{Min_2019_ICCV} employs a 3D fully-convolutional network with temporal aggregation, based on the assumption that the saliency map of any frame can be predicted by considering a limited number of past frames.

\noindent\textbf{Audiovisual saliency models:}
The first attempts in modeling audiovisual saliency were application-specific~\cite{Rue+08,schauerte_multimodal_2011,Ram+13,chen2014audio,ratajczak2016fast,sidaty2017toward}, employing mostly traditional signal processing techniques both for visual saliency and audio localization.
Coutrot and Guyader~\cite{coutrot_how_2014,coutrot_audiovisual_2014,Coutrot2016} and
Song~\cite{Song_phd13} have tried to more directly validate their
models with humans. After estimating the visual saliency map they
explicitly weigh the face image regions appropriately to generate an
audiovisual saliency map to better account for eye fixations during
movie viewing. 
Also, Min et al.~\cite{min2015fixation,min2017fixation} developed an audiovisual attention model for predicting eye fixations in scenes containing moving, sound-generating objects. For auditory attention, they employed an auditory source localization method to localize the sound-generating object on the image and fuse it with visual saliency models.
A more general model that is not application-specific that investigated ways to fuse audio and video for fixation prediction with signal processing techniques has been presented in~\cite{TSIAMI2019}. 

\noindent\textbf{Other deep audiovisual approaches:}
As mentioned before, during the last three years more and more attention is paid on integration of audio in traditionally visual-only problems, such as object localization or segmentation. In~\cite{arandjelovic2017look,arandjelovic2018objects} audiovisual correspondence in videos is learned without supervision resulting in good video and audio representations as well as in good classification in both modalities. Audiovisual representations learned in a self-supervised way and used for audiovisual scene analysis are also proposed in~\cite{senocak2018learning,owens2018audio}, while in~\cite{tian2018audio}, the authors address the problem of audiovisual event localization in unconstrained videos. A system that learns to locate image regions that produce sound and creates a representation of the sound from each pixel is described in~\cite{zhao2018sound,zhao2019sound}. Also, SoundNet~\cite{aytar2016soundnet} exploits the natural synchronization between vision and sound to learn an acoustic representation using two-million unlabeled videos, yielding improved performance in acoustic scene/object classification tasks.
In~\cite{korbar2018cooperative}, the authors exploit auditory and visual information to build effective models for both audio and video analysis from self-supervised temporal synchronization.

\section{Spatio-Temporal AudioVisual Saliency Network}
\label{sec:method}
\vspace{-0.1cm}
The proposed designed spatio-temporal audiovisual network for saliency estimation (Fig.~\ref{fig:susi}) consists of a spatio-temporal visual module that computes visual saliency, an audio representation module that computes auditory features based on~\cite{aytar2016soundnet}, a sound source localization module that computes spatio-temporal auditory saliency, an audiovisual saliency estimation module that combines and fuses the visual and auditory saliencies, and finally, the appropriate losses. These are described in detail in the next subsections.

\subsection{Spatio-Temporal Visual Network}
The architecture of our designed spatio-temporal network for visual saliency, depicted in Fig.~\ref{fig:susi}, employs the general ResNet architecture \cite{he2016deep} and specifically the 3D extension proposed initially for action classification~\cite{hara2018can}. The visual network pathway (dark purple color), with parameters $\mathbf{W}_{res}$, includes the first 4 ResNet convolutional blocks $\mathrm{conv1, conv2, conv3, conv4}$ that provide outputs $X^1, X^2, X^3, X^4$, in different spatial and temporal scales. In parallel, an attention mechanism called Deeply Supervised Attention Module (DSAM) is applied by taking the element-wise product between each channel of the feature map $X^m$ and the attention map $M^m$, to enhance the most salient regions of these feature representations:     
\begin{equation}
\tilde{X}^m = (1+M^m)\odot X^m, \quad m=1,\ldots,4.
\end{equation}
The idea of deep supervision that is the core of DSAM has been used in edge detection \cite{xie2015holistically}, object segmentation \cite{Cae+17} and static saliency \cite{wang2018deep}, but contrary to these previous works, here DSAM's role is double: It is used both for enhancing visual feature representations as well as for providing the multi-level saliency maps, as depicted by shades of green in Fig.~\ref{fig:susi}. Thus, DSAM parameters $\mathbf{W}_{am}^m$ are trained both by the main-path of the visual network and the eye-tracking data through the skip connections of the Fig.~\ref{fig:susi}. 

\begin{figure}[t]
\begin{center}
\includegraphics[width=0.45\textwidth]{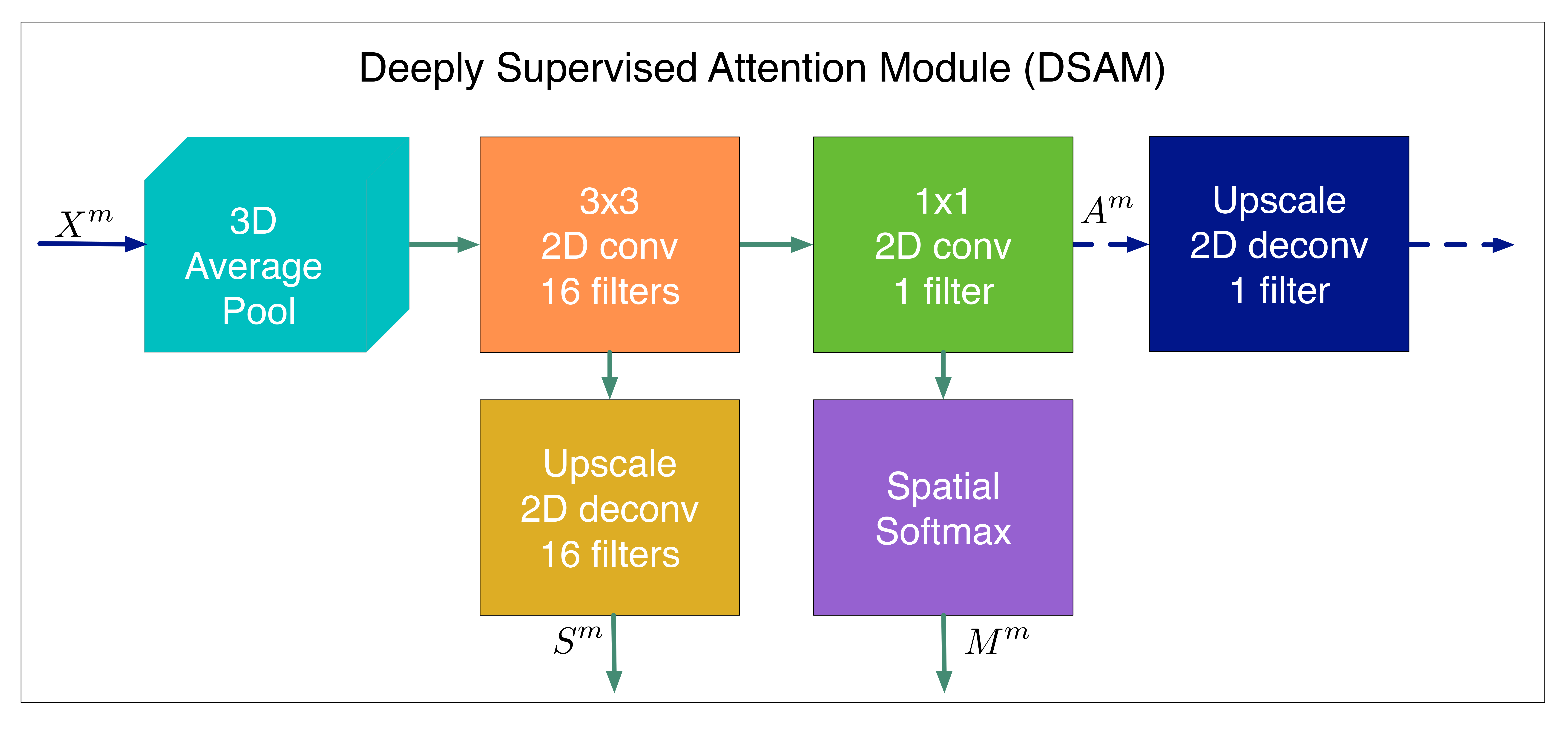}
\end{center}
\vspace{-0.3cm}
   \caption{Deeply Supervised Attention Module (DSAM) enhances the global network's representations and provides the multi-level saliency maps for spatio-temporal saliency.}
\label{fig:dsam}
\end{figure}

Figure~\ref{fig:dsam} shows the architecture of the DSAM module at level $m$. It includes an averaging pooling in the temporal dimension followed by two spatial convolution layers that provide the saliency features $S^m$ and the activation map $A^m$. Both representations are up-sampled (using the appropriate deconvolution layers) to the initial image dimensions and used for the deep supervision of the module as well as for the multi-level saliency estimation. A spatial softmax operation applied at the activation map $A^m(x,y)$ yields the attention map $M^m(x,y)$:
\begin{equation}
M^m(x,y) = \frac{\exp(A^m(x,y))}{\sum_x \sum_y \exp(A^m(x,y))}
\end{equation}

\subsection{Audio Representation Network}

For the audio representation we select to work directly on the sound waveforms and employ an 1-D fully convolutional network, rather than compute time-frequency representations and apply 2D CNNs~\cite{korbar2018cooperative,arandjelovic2017look,zhao2018sound}. First, audio is cropped in order to match the visual frames duration (i.e. 16 frames). The employed network can handle variable length audio, thus no sub-sampling is required to address the sampling rate variation from video to video. 
Subsequently, a Hanning window is applied in order to weigh higher the central audio values representing the current time instance, but also include past and future values with attenuation~\cite{TSIAMI2019}. Afterwards, for the high-level information encoding we employ a network architecture with parameters $\mathbf{W}_{a}$ based on the first seven layers of the SoundNet~\cite{aytar2016soundnet}. These layers are followed by a temporal max-pooling layer for obtaining a fixed dimension vector $f_a \in \mathbb{R}^{D_a}$ for the whole sequence. 

\subsection{Sound Source Localization in Videos}

The previous module's goal is to produce good audio representations.
Our next goal is to exploit these representations in order to detect cross-modal semantic concepts in videos. We essentially aim to retrieve correspondences between audio and spatio-temporal visual information, and thus obtain a spatio-temporal auditory saliency map. This can be viewed also as a spatial sound source localization problem. For example, a moving car, a speaking face or a rustling leaf are cross-modal events, where we need to detect where the audio comes from in the video.

Our approach selects as visual features the output of the  
3D conv3 block $X^3$ (that has feature dimension $D_v$) as in this layer with have both rich semantic information for the visual stream and quite large resolution in the spatial domain. A temporal average pooling is applied in order to marginal out the temporal dimension and have a global representation $f_v \in \mathbb{R}^{D_v \times N_X \times N_Y}$ for the whole sequence. Since visual and auditory features have different feature dimensions, we re-project both in a hidden dimension $D_h$ by applying two different affine transformations:
\begin{eqnarray}
\tilde{h}_a = \mathbf{U}_a \cdot f_a + \mathbf{b}_a, \quad
h_v = \mathbf{U}_v \cdot f_v + \mathbf{b}_v,
\end{eqnarray}
where $\tilde{h}_a \in \mathbb{R}^{D_h}$, $h_v \in \mathbb{R}^{D_h \times N_X \times N_Y}$ and $\mathbf{U}_a, \mathbf{b}_a, \mathbf{U}_v, \mathbf{b}_v$ are the corresponding learning parameters. We also apply a spatially tiling to the audio features to match the visual features spatial dimensions, getting $h_a \in \mathbb{R}^{D_h \times N_X \times N_Y}$.

For finding or learning correspondences between the audio and visual features $h_a, h_v$ we have investigated three different approaches. In the first, we simply compute the cosine similarity between the two representation vectors which does not require any learning parameters and provides a single localization map $L_1 \in \mathbb{R}^{N_X \times N_Y}$. Regarding the second approach, we take the weighted inner product of the vectors $h_a, h_v$ at each pixel $(x,y)$, and we can get either a single or multiple localization maps $L_2^j \in \mathbb{R}^{N_X \times N_Y}, \quad j=1,\dots,N_{out}$:
\begin{equation}
L_2^j(x,y) = \sum_{k=1}^{D_h} s^{j,k} \cdot h_v^k(x,y) \cdot h_a^k(x,y) + \beta^{j}, 
\end{equation}
where $s^{j,k}, \beta^{j}$ are learning parameters. In the third case, that constitutes also the proposed approach, a bilinear transformation is applied to the incoming multimodal data that can also yield either a single or multiple output maps $L_3^j \in \mathbb{R}^{N_X \times N_Y}, \quad j=1,\dots,N_{out}$:
\begin{eqnarray}
L_3^j(x,y) = h_v(x,y)^{T} \cdot \mathbf{M}^j \cdot h_a(x,y) + \mu^j \nonumber \\
= \sum_{l=1}^{D_h} \sum_{k=1}^{D_h} M^{j,l,k} \cdot h_v^l(x,y) \cdot h_a^k(x,y) + \mu^j,
\end{eqnarray}
where $M^{j,l,k}, \mu^{j}$ are the learning parameters. Note that the previous approaches ($L_1$, $L_2$) are special cases of this bilinear one ($L_3$), that allows for richer interactions between the inputs. Specifically when the matrices $\mathbf{M}^j$ are diagonal, with $s^{j,k}$ as diagonal elements, we have the case of weighted inner product ($L_2$). When the matrix $\mathbf{M}$ is the unit matrix the result is very close (modulo a normalization factor) to the cosine similarity. 


\subsection{Audiovisual Saliency Estimation}

We have now computed an auditory saliency map, expressed via source localization maps. However, there are many aspects in a video that can attract the human attention that may not be related to audio events. Thus, in order to build a multi-modal saliency estimation network that can perform well ``in-the-wild" with any kind of stimuli, we need to also include the visual-only information as modeled by the spatio-temporal visual network. For that purpose an important contribution of this paper is the investigation of different ways to perform audiovisual fusion. 

The simplest fusion scheme is to learn a linear weighted sum of the visual only map $S^v$ and an audio-related map $S^a$ (yielded by applying independent fully convolution layers to the multi-level concatenated visual saliency features $V^j~=~(S^1|\dots|S^m|\dots|S^M)$ and the localization maps $L^j$ respectively): $S_1^{av}~=~w_v~\cdot~\sigma(S^v)~+~w_a~\cdot~\sigma(S^a)$, where $\sigma(\cdot)$ denotes the sigmoid activation function. 

Moreover, inspired by previous signal processing based approaches for audiovisual saliency~\cite{TSIAMI2019,koutras2018audio}, we also investigate an attention based scheme where the audio stream modulates the visual one: $S_2^{av}~=~\sigma(S^v)~\odot~(1~+~\sigma(S^a))$. In case of multiple localization maps, we can multiply one by one the concatenated visual saliency features $V^j$ with the localization maps $L^j$ and then apply a fully convolution layer for taking the single saliency map: $\tilde{S}_2^{av,j}=\sigma(V^j) \odot (1+\sigma(L^j))$.

However, as depicted in Fig.~\ref{fig:susi}, our main and most general approach that allows more free interaction between the visual and audio-based features maps is to concatenate the multimodal features and then apply a convolution layer for their fusion in one saliency map:
$S_3^{av}~=~\mathbf{W}_{cat}~*~(V | L)~+~\bm{\beta}_{av}$.

Finally, we can also apply a late fusion scheme of all the previous approaches expressed by a weighted learned sum: $S^{av}_{fus} = \tilde{w}_{v} \cdot \sigma(S^v) + \tilde{w}_{a} \cdot \sigma(S^a) + w_{av} \cdot \sigma(S^3)$.

\subsection{Saliency Losses}
For the training of the parameters $\mathbf{W}_{v}$ that are associated with the visual stream, we construct a loss that compares the saliency map $S^v$ and the activations $A^m$ with the ground truth maps $Y_{sal}$ obtained by the eye-tracking data:
\begin{equation}
\begin{split}
\mathcal{L}_{v}(\mathbf{W}_{v}) = \mathcal{D}(\mathbf{W}_{v} | \sigma(S^v),Y_{sal}) + \\ 
\sum_{m=1}^4 \mathcal{D}(\mathbf{W}_{AM}^m | \sigma(A^m),Y_{sal}), \label{sal_loss_gen_v}
\end{split}
\end{equation}
where $\sigma(\cdot)$ denotes the sigmoid non-linearity and $\mathcal{D}(\cdot)$ is a loss function between the estimated and the ground truth 2D maps. 
When we train the parameters $\mathbf{W}_{av}$ of the audiovisual network, we use the trained visual-only network as a starting point and we do not use the skip connections of the DSAM modules:
\begin{equation}
\mathcal{L}_{av}(\mathbf{W}_{av}) = \mathcal{D}(\mathbf{W}_{av} | \sigma(S^{av}),Y_{sal}). \label{sal_loss_gen_av}
\end{equation}

Several different metrics to evaluate saliency are employed in order to compare the predicted saliency map $P \in [0,1]^{N_X \times N_Y}$ to the eye-tracking data~\cite{bylinskii2016different}. As ground truth maps we can use either the map of fixation locations $Y_{fix} \in \{0,1\}^{N_X \times N_Y}$ on the image plane of size $N_X \times N_Y$ or the dense saliency map $Y_{den}\in[0,1]^{N_X \times N_Y}$ that is produced by convolving the binary fixation map with a gaussian kernel. Thus, as $\mathcal{D}(\cdot)$ we employ three loss functions associated with the different aspects of saliency evaluation. The first is the cross-entropy loss between the predicted map $P$ and the dense map $Y_{den}$:
\begin{equation}
\begin{split}
\mathcal{D}_{CE}(\mathbf{W}|P,Y_{den})= - \sum_{x,y} Y_{den}(x,y) \odot \log(P(x,y;\mathbf{W})) \\
+ (1-Y_{den}(x,y)) \odot (1-\log(P(x,y;\mathbf{W}))).
\end{split}
\end{equation}

\noindent The second loss function is based on the linear Correlation Coefficient (CC)~\cite{bylinskii2016different} that is widely used in saliency evaluation and measures the linear relationship between the predicted saliency $P$ and the dense ground truth map $Y_{den}$:
\begin{equation}
\mathcal{D}_{CC}(\mathbf{W}|P,Y_{den}) = -\frac{\mathrm{cov}(P(x,y;\mathbf{W}),Y_{den}(x,y))}{\rho(P(x,y;\mathbf{W}))\cdot\rho(Y_{den}(x,y))}, 
\end{equation} 
where $\mathrm{cov},\rho$ denote the covariance and the standard deviation respectively. 
The last loss is derived from the Normalized Scanpath Saliency (NSS) metric~\cite{bylinskii2016different}, which is computed
as the estimated map values $\tilde{P}(x,y;\mathbf{W})=\frac{P(x,y;\mathbf{W})-\mu(P(x,y;\mathbf{W}))}{\rho(P(x,y;\mathbf{W}))}$, after zero mean normalization and unit standardization, at human fixation locations ($Y_{fix}(x,y)=1$):
\begin{equation}
\mathcal{D}_{NSS}(\mathbf{W}|\tilde{P},Y_{fix}) = - \frac{1}{N_{f}} \sum_{x,y} \tilde{P}(x,y;\mathbf{W})\odot Y_{fix}(x,y),
\end{equation}
where $N_{f}=\sum_{x,y}Y_{fix}(x,y)$ denotes the total number of fixation points. 
The final loss of the $i$-th input sample is given by a weight combination of the losses $\mathcal{L}^i_{CE}, \mathcal{L}^i_{CC}, \mathcal{L}^i_{NSS}$ that are given either by Eq.(\ref{sal_loss_gen_v}) or Eq.(\ref{sal_loss_gen_av}) using the corresponding loss functions $\mathcal{D}^i_{CE}, \mathcal{D}^i_{CC}, \mathcal{D}^i_{NSS}$:
\begin{equation}
\mathcal{L}_{sal}^i(\mathbf{W}) = w_{1}\mathcal{L}^i_{CE} + w_{2}\mathcal{L}^i_{CC} + w_{3}\mathcal{L}_{NSS}^i,
\end{equation}  
where $w_1,w_2,w_3$ are the weights of each loss type.

\subsection{Implementation}

Our implementation and experimentation with the visual network uses as backbone the 3D ResNet-50 architecture \cite{hara2018can} that has showed competitive performance against other deeper architectures for action recognition task, in terms of performance and computational budget. As starting point for $\mathbf{W}_{res}$ we used the weights from the pretrained model in the Kinetics 400 database. For the audio stream we employ the SoundNet (using 7 out of 8 layers) architecture~\cite{aytar2016soundnet} that is based on 1D temporal convolution and has been successfully applied in acoustic scene/event classification tasks. As starting point for our audio representation network parameters $\mathbf{W}_a$ we use the weights from the pretrained model that has been trained on two million videos from Flickr~\cite{aytar2016soundnet}.
\noindent \textbf{Training:} For training we employ stochastic gradient descent with momentum 0.9, while we assign a weight decay of 1e-5 for regularization. We have also employed effective batchsizes of 128 samples, and multistep learning rate. First we train the visual-only spatio-temporal network and afterwards the whole audiovisual network using the  visual network's weights as starting points. Note that DSAM modules' skip connections are used only for visual-only network training.
The weights $w_1, w_2, w_3$ for the saliency loss are selected as 0.1, 2, 1 respectively, after experimentation.

\begin{figure}[t]
\begin{center}
\includegraphics[width=\linewidth]{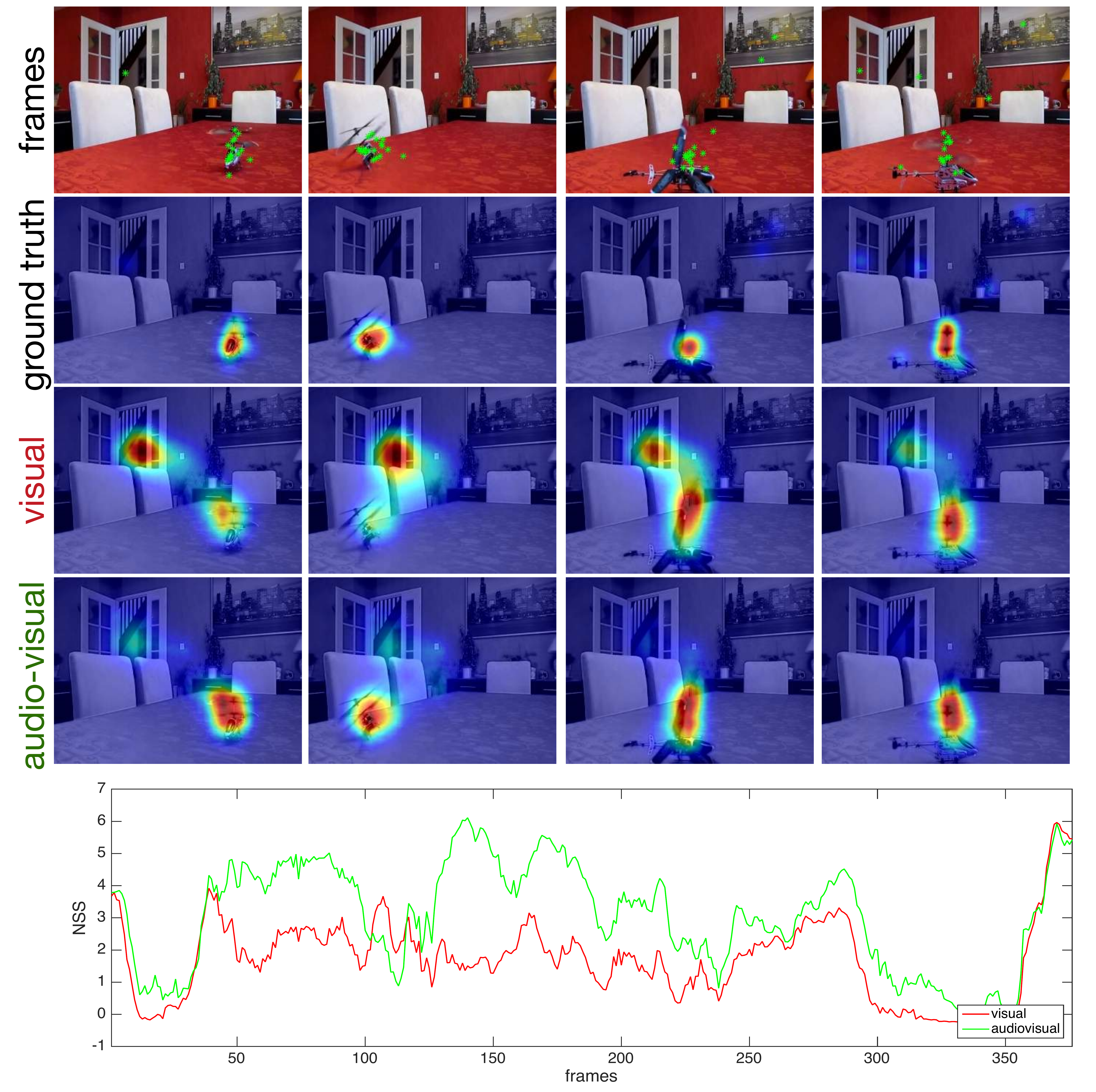}
\end{center}
\vspace{-0.4cm}
   \caption{Sample frames from Coutrot1 database with their eye-tracking data, and the corresponding ground truth, spatio-temporal visual, and audiovisual saliency maps as produced by STAViS (visual-only and audiovisual). Also NSS curve over time for visual and audiovisual approaches is depicted.}
\label{fig:coutrot1}
\vspace{-0.3cm}
\end{figure}

\begin{figure}[t]
\begin{center}
\includegraphics[width=\linewidth]{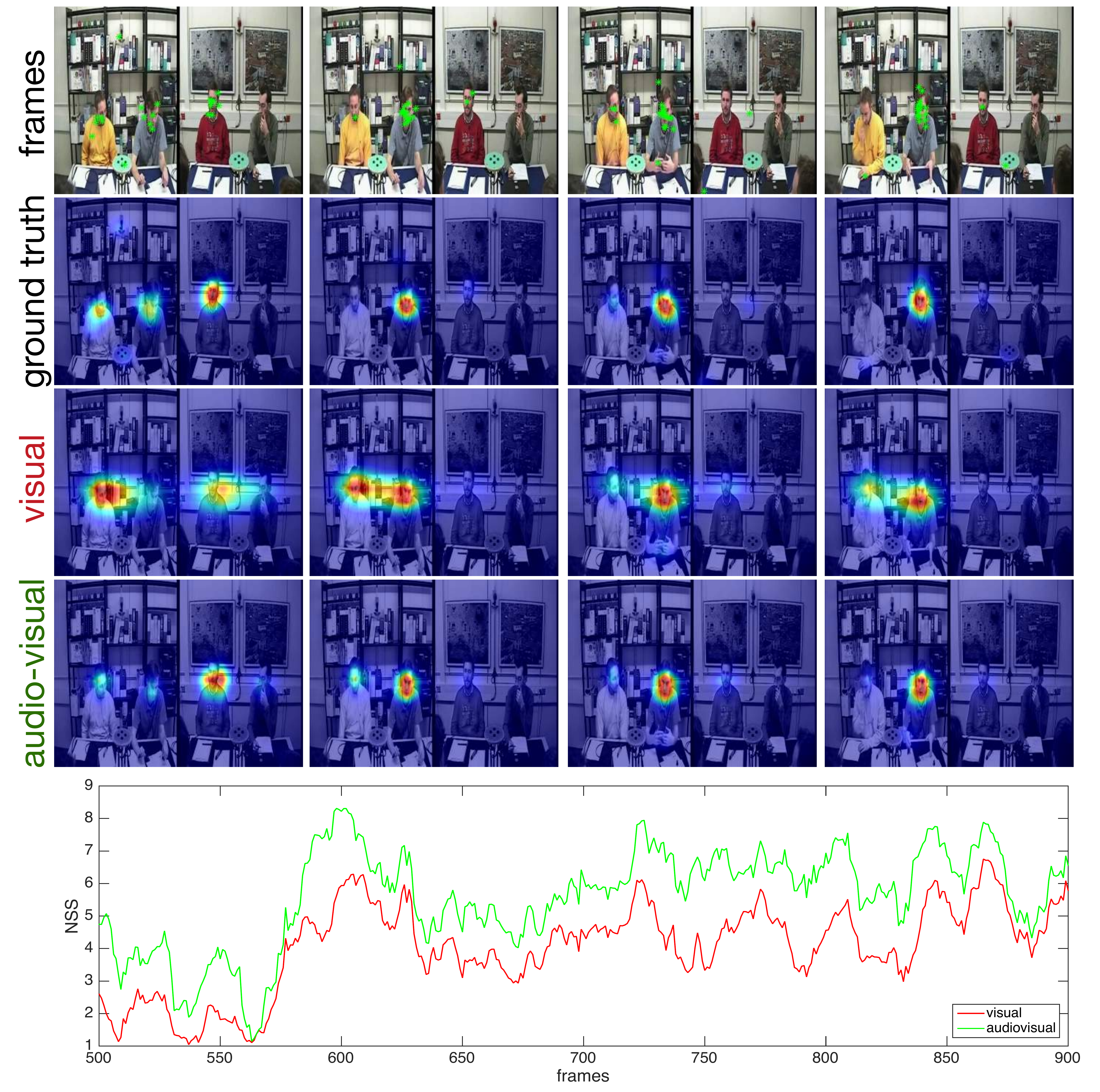}
\end{center}
\vspace{-0.4cm}
   \caption{Sample frames from Coutrot2 database with their eye-tracking data, and the corresponding ground truth, spatio-temporal visual, and audiovisual saliency maps as produced by STAViS (visual-only and audiovisual). Also NSS curve for visual and audiovisual approaches is depicted.}
\label{fig:coutrot2}
\vspace{-0.3cm}
\end{figure}

\noindent \textbf{Data Augmentation:} The input samples in the network consist of 16-frames RGB video clips spatially resized at $112 \times 112$ pixels, as well as the corresponding audio stream.
We have also applied data augmentation for random generation of training samples. First, we divided the initial long-duration videos into 90-frame, non-overlapping segments and then generated the 16-frames from each segment by applying a random flipping process, as in \cite{wang2016temporal}, but without random cropping. We applied the same spatial transformations to the 16 frames of the video clip and the eye-tracking based saliency map of the median frame, which is considered the clip's  ground truth map.

\noindent \textbf{Testing:} During testing phase, we obtain an estimated saliency map per frame using a 16-frame sliding window with step 1 without any random flipping.     

\section{Experiments}

\begin{table}[t]
 \centering{
 \resizebox{0.48\textwidth}{!}{
 \begin{tabular}{ | l | c c c c c | }
 \hline
 \multirow{2}{*}{\backslashbox{ \kern-0.5em Method \kern-0.5em}{\kern-1.9em Dataset \kern-0.5em}} &  \multicolumn{5}{c|}{Overall}  \\ \cline{2-6} 
 & CC $\uparrow$ & NSS $\uparrow$ & AUC-J $\uparrow$ & sAUC $\uparrow$ & SIM  $\uparrow$ \\ \hline \hline
Visual & 0.5260 & 2.54 & 0.8922 & 0.6187 & 0.4088  \\ \hline
$L_1$\_AudioOnly & 0.5132	& 2.48 & 0.8923 & 0.6220 & 0.4100 \\ 
$L_2$\_AudioOnly & 0.5381 & 2.63 & 0.8969 & 0.6171 & 0.4157\\ 
$L_3$\_AudioOnly & 0.5226 & 2.59 & 0.8935 & 0.6228 & 0.4027 \\ \hline
$L_1$\_$S_1^{av}$ & 0.5243 & 2.62 & 0.8924 & 0.6224 & 0.4103  \\ 
$L_2$\_$S_1^{av}$ & 0.5344 & 2.69 & 0.8944 & 0.6215 & 0.4249 \\ 
$L_3$\_$S_1^{av}$ & 0.5354 & 2.71& 0.8959 & 0.6236 & 0.4278 \\ \hline
$L_1$\_$S_2^{av}$ & 0.5069 & 2.43 & 0.8687 & 0.6223 & 0.2985 \\ 
$L_2$\_$S_2^{av}$ & 0.5082 & 2.44 & 0.8699 & 0.6225 & 0.2996 \\ 
$L_3$\_$S_2^{av}$ & 0.5066 & 2.46 & 0.8622 & 0.6219 & 0.2981 \\  
$L_3^{mul}$\_$\tilde{S}_2^{av}$ & 0.5271 & 2.56 & 0.8943 & 0.6242 & 0.4132 \\ \hline
$\mathbf{L_3^{mul}}$\_$\mathbf{S_3^{av}}$ & \textbf{0.5414} & \textbf{2.73} & \textbf{0.8983} & \textbf{0.6267} & 0.4241 \\ \hline
$L_3^{mul}$\_$S_{fus}^{av}$ & 0.5401 & 2.70 & 0.8979 & 0.6261 & \textbf{0.4290} \\ \hline
\end{tabular}
}}
\vspace{0.1cm}
\caption{Ablation study: $L_1, L_2$ and $L_3$ refer to the audio localization method: cosine, inner product and bilinear respectively. $S_1^{av}, S_2^{av}$, $S_3^{av}$ and $S_{fus}^{av}$ refer to the different fusion methods. Superscript ``mul" refers to multiple feature maps.}
\vspace{-0.5cm}
\label{table:sal_eval_ablation}
\end{table}

\begin{table*}[th!]
 \centering{
 \resizebox{0.98\textwidth}{!}{
 \begin{tabular}{ l | c c c c c || c c c c c || c c c c c}
 \hline
 \multirow{2}{*}{\backslashbox{ \kern-0.5em Method \kern-0.5em}{\kern-1.9em Dataset \kern-0.5em}} &  \multicolumn{5}{c||}{DIEM}   &  \multicolumn{5}{c||}{Coutrot1} & \multicolumn{5}{c}{Coutrot2} \\ \cline{2-16} 
 & CC $\uparrow$ & NSS $\uparrow$ & AUC-J $\uparrow$ & sAUC $\uparrow$ & SIM  $\uparrow$ & CC $\uparrow$ & NSS $\uparrow$ & AUC-J $\uparrow$ & sAUC $\uparrow$ & SIM  $\uparrow$ & CC $\uparrow$ & NSS $\uparrow$ & AUC-J $\uparrow$ & sAUC $\uparrow$ & SIM  $\uparrow$\\ \hline \hline
\textbf{STAViS} [STA] & \textbf{0.5795} & \textbf{2.26} & \textbf{0.8838} & \textbf{0.6741} & \textbf{0.4824} & 
0.4722 & 2.11 & \textbf{0.8686} & \textbf{0.5847} & \textbf{0.3935} &
\textbf{0.7349} & \textbf{5.28} & \textbf{0.9581} & \textbf{0.7106} & \textbf{0.5111} \\ \hline \hline 
STAViS [ST] & 0.5665 & 2.19 & 0.8792 & 0.6648 & 0.4719 & 0.4587 & 1.99 & 0.8617 & 0.5764 & 0.3842 &
0.6529 & 4.19 & 0.9405 & 0.6895 & 0.4470 \\ 
DeepNet \cite{Pan_2016_CVPR} [S]& 0.4075 & 1.52 & 0.8321 & 0.6227 & 0.3183& 
0.3402 & 1.41 & 0.8248 & 0.5597 & 0.2732 & 
0.3012 & 1.82 & 0.8966 & 0.6000 & 0.2019 \\
DVA \cite{wang2018deep} [S] & 0.4779 & 1.97 & 0.8547 & 0.641 & 0.3785 & 
0.4306 & 2.07 & 0.8531 & 0.5783 & 0.3324
& 0.4634 & 3.45 & 0.9328 & 0.6324 & 0.2742 \\
SAM \cite{cornia2018predicting} [S] & 0.4930 & 2.05 & 0.8592 & 0.6446 & 0.4261 &
0.4329 & 2.11 & 0.8571 & 0.5768 & 0.3672 & 
0.4194 & 3.02 & 0.9320 & 0.6152 & 0.3041 \\ 
SalGAN \cite{Pan_2017_SalGAN} [S] & 0.4868 & 1.89 & 0.8570 & 0.6609 & 0.3931 &
0.4161 & 1.85 & 0.8536 & 0.5799 & 0.3321 & 
0.4398 & 2.96 & 0.9331 & 0.6183 & 0.2909 \\
ACLNet \cite{wang2018revisiting,wang2019revisiting} [ST] & 0.5229 & 2.02 & 0.8690 & 0.6221 & 0.4279 & 
0.4253 & 1.92 & 0.8502 & 0.5429 & 0.3612 & 
0.4485 & 3.16 & 0.9267 & 0.5943 & 0.3229 \\
DeepVS \cite{jiang2018deepvs} [ST] & 0.4523 & 1.86 & 0.8406 & 0.6256 & 0.3923 &
0.3595 & 1.77 & 0.8306 & 0.5617 & 0.3174 & 
0.4494 & 3.79 & 0.9255 & 0.6469 & 0.2590 \\ 
TASED \cite{Min_2019_ICCV} [ST] & 0.5579 & 2.16 & 0.8812 & 0.6579 & 0.4615 &
\textbf{0.4799} & \textbf{2.18} & 0.8676 & 0.5808 & 0.3884 & 
0.4375 & 3.17 & 0.9216 & 0.6118 & 0.3142 \\ \hline 
\end{tabular}
}}
\vspace{0.1cm}
\caption{Evaluation results for saliency in DIEM, Coutrot1 and Coutrot2 databases. The proposed method's (STAViS [STA]) results are depicted in the first row, while the second one refers to our visual-only version. In most cases, the proposed network outperforms the existing state-of-the-art methods for saliency estimation according the five evaluation metrics. [STA] stands for spatio-temporal audiovisual, [ST] for spatio-temporal visual models while [S] denotes a spatial only model that is applied to each frame independently.}
\label{table:sal_eval}
\end{table*}

\begin{table*}[th!]
 \centering{
 \resizebox{0.98\textwidth}{!}{
 \begin{tabular}{ l | c c c c c || c c c c c || c c c c c}
 \hline
 \multirow{2}{*}{\backslashbox{ \kern-0.5em Method \kern-0.5em}{\kern-1.9em Dataset \kern-0.5em}} &  \multicolumn{5}{c||}{AVAD}   &  \multicolumn{5}{c||}{SumMe} & \multicolumn{5}{c}{ETMD} \\ \cline{2-16} 
 & CC $\uparrow$ & NSS $\uparrow$ & AUC-J $\uparrow$ & sAUC $\uparrow$ & SIM  $\uparrow$ & CC $\uparrow$ & NSS $\uparrow$ & AUC-J $\uparrow$ & sAUC $\uparrow$ & SIM  $\uparrow$ & CC $\uparrow$ & NSS $\uparrow$ & AUC-J $\uparrow$ & sAUC $\uparrow$ & SIM  $\uparrow$\\ \hline \hline
\textbf{STAViS} [STA] & \textbf{0.6086} & \textbf{3.18} & \textbf{0.9196} & \textbf{0.5936} & \textbf{0.4578} &
0.4220 & 2.04 & \textbf{0.8883} & 0.6562 & \textbf{0.3373} &
\textbf{0.5690} & \textbf{2.94} & \textbf{0.9316} & 0.7317 & \textbf{0.4251} \\ \hline \hline 
STAViS [ST] & 0.6041 & 3.07 & 0.9157 & 0.5900 & 0.4431 & 0.4180 & 1.98 & 0.8848 & 0.6477 & 0.3325 &
0.5602 & 2.84 & 0.9290 & 0.7278 & 0.4121 \\ 
DeepNet \cite{Pan_2016_CVPR} [S]& 0.3831 & 1.85 & 0.8690 & 0.5616 & 0.2564& 
0.3320 & 1.55 & 0.8488 & 0.6451 & 0.2274 & 
0.3879 & 1.90 & 0.8897 & 0.6992 & 0.2253 \\
DVA \cite{wang2018deep} [S] & 0.5247 & 3.00 & 0.8887 & 0.5820 & 0.3633 & 
0.3983 & 2.14 & 0.8681 & 0.6686 & 0.2811
& 0.4965 & 2.72 & 0.9039 & 0.7288 & 0.3165 \\
SAM \cite{cornia2018predicting} [S] & 0.5279 & 2.99 & 0.9025 & 0.5777 & 0.4244 &
0.4041 & \textbf{2.21} & 0.8717 & 0.6728 & 0.3272 & 
0.5068 & 2.78 & 0.9073 & 0.7310 & 0.3790 \\ 
SalGAN \cite{Pan_2017_SalGAN} [S] & 0.4912 & 2.55 & 0.8865 & 0.5799 & 0.3608 &
0.3978 & 1.97 & 0.8754 & \textbf{0.6882} & 0.2897 & 
0.4765 & 2.46 & 0.9035 & \textbf{0.7463} & 0.3117 \\ 
ACLNet \cite{wang2018revisiting,wang2019revisiting} [ST] & 0.5809 & 3.17 & 0.9053 & 0.5600 & 0.4463 & 
0.3795 & 1.79 & 0.8687 & 0.6092 & 0.2965 & 
0.4771 & 2.36 & 0.9152 & 0.6752 & 0.3290 \\
DeepVS \cite{jiang2018deepvs} [ST] & 0.5281 & 3.01 & 0.8968 & 0.5858 & 0.3914 &
0.3172 & 1.62 & 0.8422 & 0.6120 & 0.2622 & 
0.4616 & 2.48 & 0.9041 & 0.6861 & 0.3495 \\ 
TASED \cite{Min_2019_ICCV} [ST] & 0.6006 & 3.16 & 0.9146 & 0.5898 & 0.4395 &
\textbf{0.4288} & 2.10 & 0.8840 & 0.6570 & 0.3337 & 
0.5093 & 2.63 & 0.9164 & 0.7117 & 0.3660 \\ \hline 
\end{tabular}
}}
\vspace{0.1cm}
\caption{Evaluation results for audiovisual saliency in AVAD, SumMe and ETMD databases. The proposed method's (STAViS [STA]) results are depicted in the first row, while the second one refers to our visual-only version. In most cases, the proposed network outperforms the existing state-of-the-art methods for saliency estimation according the five evaluation metrics. [STA] stands for spatio-temporal audiovisual, [ST] for spatio-temporal visual models while [S] denotes a spatial only model that is applied to each frame independently.}
\vspace{-0.2cm}
\label{table:sal_eval2}
\end{table*}

\subsection{Datasets}

In order to train and evaluate the proposed audiovisual saliency network, 6 different datasets are employed, containing audiovisual eye-tracking data: DIEM, AVAD, Coutrot1, Coutrot2, SumMe, and ETMD. These databases consist of various types of videos, ranging from very structured small videos to completely unstructured, user-made Youtube videos. Our goal is to train and evaluate the model using different types of audiovisual data, in order to obtain a good performance ``in-the-wild" and help the model learn if and when it should use audio to enhance visual saliency. 
A short description for each database follows.

\noindent\textbf{AVAD:}
AVAD database~\cite{min2017fixation} contains 45 short clips of 5-10 sec duration with several audiovisual scenes, e.g. dancing, guitar playing, bird signing, etc. 
A joint audiovisual event per video is always present. 
Eye-tracking data from 16 participants have been recorded.

\noindent\textbf{Coutrot databases:}
Coutrot databases~\cite{coutrot_how_2014,Coutrot2016} are split in Coutrot1 and Coutrot2: Coutrot1 contains 60 clips with dynamic natural scenes split in 4 visual categories: one/several moving objects, landscapes, and faces. Eye-tracking data from 72 participants have been recorded. Coutrot2 contains 15 clips of 4 persons in a meeting and the corresponding eye-tracking data from 40 persons.

\noindent\textbf{DIEM:}
DIEM database~\cite{mital+2011} consists of $84$ movies of all sorts, sourced from
publicly  accessible  repositories, including
advertisements, documentaries, game trailers, movie trailers, music videos, news clips, and time-lapse footage. Thus, the majority of DIEM videos are documentary-like, which means that audio and visual information do not correspond to the same event. Eye movement data from $42$ participants were recorded via an Eyelink eye-tracker, while watching the videos in random order and with the audio on.

\noindent\textbf{SumMe:}
SumMe database~\cite{gygli+2014, TSIAMI2019} contains $25$ unstructured videos, i.e. mostly user-made videos, as well as their corresponding multiple-human created summaries, which were acquired in a controlled psychological experiment. 
Audiovisual eye-tracking data have been collected~\cite{TSIAMI2019} from $10$ viewers, recorded via an Eyelink eye-tracker.

\noindent\textbf{ETMD:}
ETMD database~\cite{koutras2015,TSIAMI2019} contains $12$ videos from six different hollywood movies. Audiovisual eye-tracking data have been collected~\cite{TSIAMI2019} from $10$ viewers, recorded via an Eyelink eye-tracker.

\subsection{Experimental Results}

Training has been performed by combining data from all datasets: For DIEM, the standard split from literature has been employed~\cite{borji+13}. For the other 5 databases, where there is no particular split, we created 3 different splits of the data, in the sense of 3-fold cross-validation, with non-overlapping videos between train, validation and test sets for each split, uniformly split among datasets. Among these 3 different splits, different videos were placed in each split, namely, if video1 of dataset1 was placed in test set of split1, then it would not appear in any other test set. We ensured that all videos of each dataset would appear once in the test set of one split. 
The models' final performance was obtained by taking the average among all 3 splits.
The same procedure was carried out both for our audiovisual and visual variants, in order to ensure  a fair comparison.

For the evaluation of STAViS network, we first investigate the performance of the various sound localization and fusion techniques described in Sec.~\ref{sec:method}. We  pick the method with the best results and compare it to 8 state-of-the-art visual saliency methods (using their publicly available codes and models), in all 6 databases on the same test data. We employ five widely-used evaluation metrics for saliency \cite{bylinskii2016different}: CC, NSS, AUC-Judd (AUC-J), shuffled AUC (sAUC) and SIM (similarity). For sAUC we select the negative samples from the union of all viewers' fixations across all other frames except the frame for which we compute the AUC.

\noindent~\textbf{Ablation study:}
Regarding Table~\ref{table:sal_eval_ablation}, we can observe that bilinear transformation $L_3$ achieves better performance compared to the other sound localization methods since the latter can be consider as simpler versions of the bilinear function. Moreover, it can be noted that methods based only on audio localization can achieve quite high scores while they are further improved when fused with pure visual cues.
Especially when multiple maps are fused with the visual ones by concatenation $S_3^{av}$, thus allowing for the most interaction between the features, the best performance is achieved in almost all metrics.  Competitive performance is also achieved with late fusion, as indicated by the last row of Table~\ref{table:sal_eval_ablation}. Overall, cosine similarity for localization $L_1$ performs worst in all cases, since it does not have any learnable parameters, while fusion scheme $S_2^{av}$ that has performed well in signal processing based methods~\cite{TSIAMI2019} does not seem suitable for deep learning approaches. Thus, we pick up method $L_3^{mul} S_3^{av}$ as our proposed STAViS method. Figures~\ref{fig:coutrot1}, \ref{fig:coutrot2} depict some sample frames from Coutrot1 and Coutrot2 along with their corresponding eye-tracking data and the ground truth, visual-only (ours) and the  STAViS saliency maps. Attention is much better captured by the proposed STAViS model. They also depict NSS curve over time for visualization and comparison purposes.

\noindent~\textbf{Comparisons with state-of-the-art:} 
Next, we proceed with extensive comparisons with 8 different state-of-the-art saliency methods, depicted for the five metrics per database, in Tables~\ref{table:sal_eval},~\ref{table:sal_eval2}. Results highlight the superiority of our proposed audiovisual method, as it outperforms almost for all databases and metrics the other state-of-the-art methods. We can observe that all the other state-of-the-art methods, except for TASED, achieve good performance in some databases, but not consistently, meaning that they might perform well for particular types of video but not for all types. For example, ACLNet performs well for AVAD database but not equally well in SumMe which has more unstructured, home-made videos. TASED achieves good performance, even better than STAViS for CC and NSS in Coutrot1 and SumMe, but it does not perform equally well for Coutrot2. Overall, a good performance is achieved from our visual-only version as well.

\begin{figure}[t]
\begin{center}
\includegraphics[width=\linewidth]{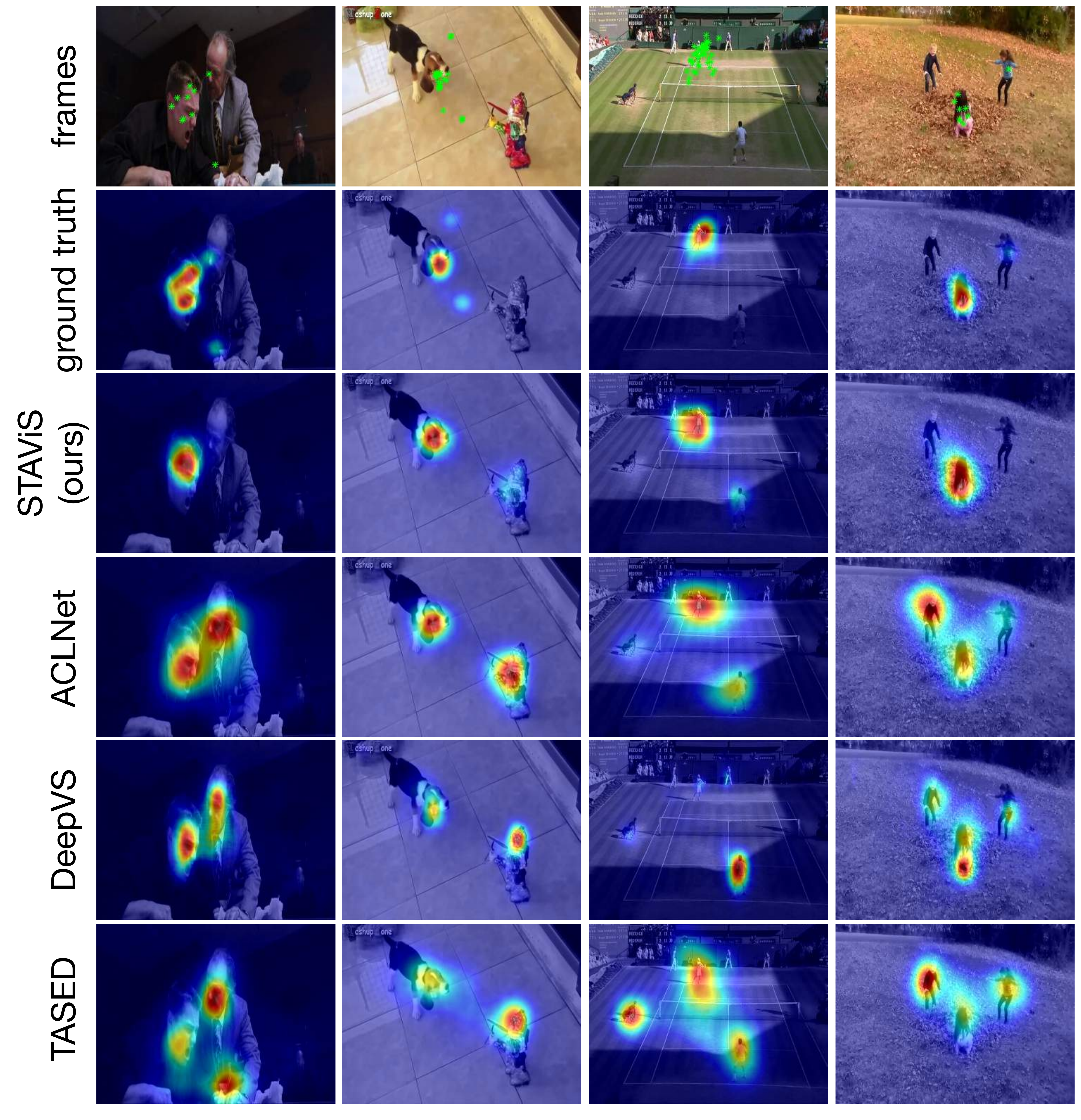}
\end{center}
\vspace{-0.4cm}
   \caption{Sample frame from ETMD, AVAD, DIEM and SumMe databases with their eye-tracking data, and the corresponding ground truth, STAViS, and other spatio-temporal state-of-the-art visual saliency maps for comparisons.}
\label{fig:sota_sample_frames}
\vspace{-0.3cm}
\end{figure}

\noindent\textbf{Discussion:} Our proposed spatio-temporal auditory saliency model STAViS performs consistently well across all datasets, achieving the best performance compared to state-of-the-art methods for almost all databases and metrics. It should be noted that these databases contain a large variety of videos types. From documentaries, where there is no audiovisual correspondence, to Hollywood movies and unprocessed user-made videos. Thus our model achieves a good performance ``in-the-wild", without any prior knowledge on the content of the video. This performance for this large range could indicate that our model is capable of learning if, where and when auditory saliency integration should be performed. 
In addition, especially in Coutrot2 that is the most ``audiovisual" database in the sense that all included events/videos consist of audiovisual scenes, our model clearly outperforms the rest by a large margin probably because it better captures the audiovisual correspondences. 
For visualization purposes, in Fig.~\ref{fig:sota_sample_frames} a sample frame from ETMD, AVAD, DIEM and SumMe database is presented with its corresponding eye-tracking data in the first row. The second row depicts the ground truth saliency map. The third row includes the corresponding saliency maps from our STAViS model, while the final 3 rows, the same maps for ACLNet, DeepVS and TASED. It can easily be observed that our results are closer to the ground truth.

\section{Conclusions}

We have proposed STAViS, a novel spatio-temporal audiovisual network that efficiently addresses the problem of fixation prediction in videos, i.e. video saliency estimation. It employs an extended state-of-the-art visual saliency network, a state-of-the-art audio representation network and features a sound source localization module that produces one or multiple auditory saliency maps and a fusion module that combines auditory and visual saliency maps in order to produce an audiovisual saliency map. All components have been designed, trained end-to-end, and extensively evaluated in a variety of videos. Results for 5 different metrics in 6 different databases and comparison with 8 state-of-the-art methods indicate the appropriateness and efficiency of this audiovisual approach for saliency modeling.


{\small
\bibliographystyle{ieee_fullname}
\bibliography{strings,egbib}
}

\section{Supplementary material}

\subsection{Configuration and parameters of the employed CNN architectures}

\begin{figure}[h]
\begin{center}
\includegraphics[width=\linewidth]{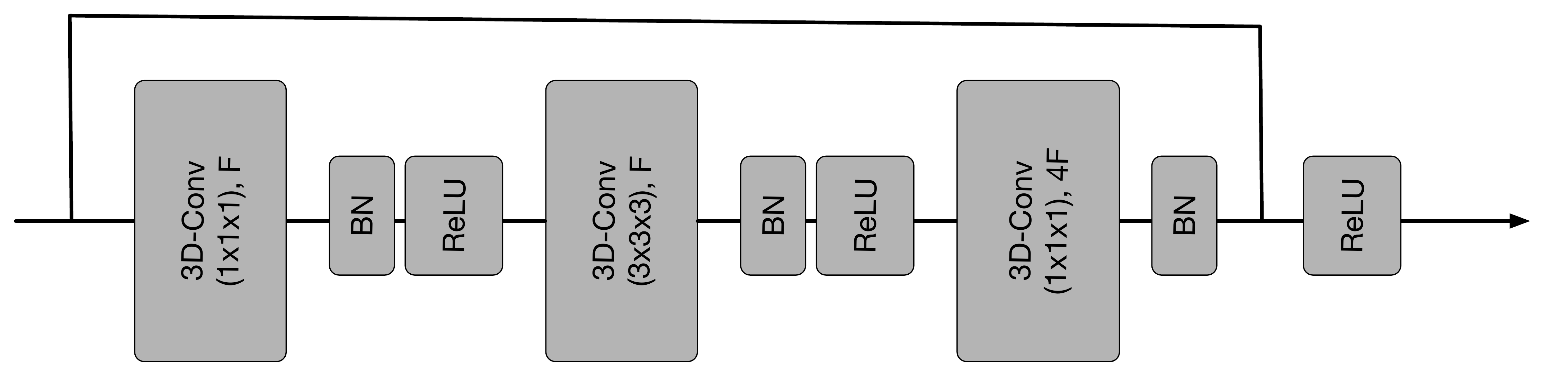}
\end{center}
\vspace{-0.4cm}
   \caption{Architecture of the 3D Bottleneck Block. $F$ denotes the number of feature maps of the 3D convolutional filter while BN refers to batch normalization.}
\label{fig:bottle}
\vspace{-0.3cm}
\end{figure}

In Tables~\ref{table:resnet}, \ref{table:soundnet} we present the configuration and the parameters of the employed visual and audio sub-networks based on the 3D-ResNet-50 \cite{hara2018can} and SounNet \cite{aytar2016soundnet} architectures respectively. Figure~\ref{fig:bottle} depicts the architecture of a 3D Bottleneck Block that constitutes the main building unit of the spatio-temporal 3D ResNet.

\begin{table*}[h]
 \centering{
 \begin{tabular}{ l | c | c | c | c | c  }
 Layer & 3D-Conv1 & 3D Max Pool & 3D Conv2 Block & 3D Conv3 Block & 3D Conv4 Block \\ \hline 
Number of feature maps ($F$) & 64 & 64 & 64 & 128 & 256 \\
Filter Kernel & $7\times7\times7$ & $3\times3\times3$ & Bottleneck & Bottleneck & Bottleneck \\
 Stride & $7\times7\times1$  & $2\times2\times2$ & $1\times1\times1$ & $2\times2\times2$ & $2\times2\times2$ \\
Number of Blocks & - & - & 3 & 4 & 6 \\
 \end{tabular}
}
\vspace{0.3cm}
\caption{Configuration and parameters of the 3D ResNet architecture that has been employed in the patio-temporal network for visual saliency.}
\label{table:resnet}
\end{table*}

\begin{table*}[h]
 \centering{
 \begin{tabular}{ l | c c | c c | c | c | c c | c | c }
 Layer & Conv1 & Pool1 & Conv2 & Pool2 & Conv3 & Conv4 & Conv5 & Pool5 & Conv6 & Conv7 \\ \hline 
 Number of feature channels & 16 & 16 & 32 & 32 & 64 & 128 & 256  & 256 & 512 & 1024 \\
 Filter Kernel & 64 & 8 & 32 & 8 & 16 & 8 & 4 & 4 & 4 & 4 \\
 Stride & 2 & 1 & 2 & 1 & 2 & 2 & 2 & 1 & 2 & 2 \\
 Padding & 32 & 0 & 16 & 0 & 8& 4 & 2 &0 & 2&2
 \end{tabular}
}
\vspace{0.3cm}
\caption{Configuration and parameters of the Audio Representation Network.}
\label{table:soundnet}
\end{table*}

\subsection{Data splits and training}

\subsubsection{Data splits}
As mentioned in the main body of the paper, training has been performed by combining data from all datasets: For DIEM, the standard split from literature has been employed~\cite{borji+13}. For the other 5 databases, where there is no particular split, we created 3 different splits of the data, in the sense of 3-fold cross-validation, with non-overlapping videos between train, validation and test sets for each split, uniformly split among datasets. Among these 3 different splits, different videos were placed in each split, namely, if video1 of dataset1 was placed in test set of split1, then it would not appear in any other test set. We ensured that all videos of each dataset would appear once in the test set of one split. 
The models' final performance was obtained by taking the average among all 3 splits.
The same procedure was carried out both for our audiovisual and visual variants, in order to ensure  a fair comparison.
In Table~\ref{table:splits}, the detailed list of the video contents of each split is depicted for completeness.

\subsubsection{Training}
In this subsection, the training process is described in more detail. First the visual-only network is trained, using as starting point the pretrained model in the Kinetics 400 database \cite{hara2018can,carreira2017quo} and employing DSAM skip connections. Afterwards, the whole audio-visual network is trained end-to-end, using the previously trained visual-only network as starting point for the visual path, while for the audio representation path we use as starting point the pretrained model from Flickr \cite{aytar2016soundnet}. The network was trained either for 100 epochs or until loss did not further improve during 10 epochs, which usually happened around 60 epochs. Also, multi-step learning rate has been employed. The hyperparameters listed in the main body of the paper have mainly been decided upon experimentation. 

\begin{table*}[t!]
 \centering{
 \begin{tabular}{ | c | c | c | c | }
 \hline
& Test split 1 & Test split 2 & Test split 3 \\ \hline
 Database & video & video & video \\ \hline
AVAD & V32\_Dancers & V22\_Tap2 & V40\_Guitar5\\ 
AVAD & V6\_Basketball1 & V34\_Beat & V16\_Drummer2\\ 
AVAD & V33\_Harp  & V35\_Squirrel & V30\_Dog3\\  
AVAD & V17\_Soccer1 & V3\_Speech3 & V7\_Basketball2\\  
AVAD & V41\_Violin1 & V11\_News4 & V42\_Violin2\\  
AVAD & V29\_Dog2 & V45\_Darbuka2 & V28\_Dog1\\  
AVAD & V25\_Piano1 & V38\_Guitar3 & V5\_Interview2\\  
AVAD & V31\_Bird & V14\_Conservation3 & V21\_Tap1\\  
AVAD & V27\_Piano3 & V8\_News1 & V37\_Guitar2\\  
AVAD & V15\_Drummer1 & V44\_Darbuka1 & V26\_Piano2\\  
AVAD & V23\_Tap3 & V1\_Speech1 & V43\_Violin3\\  
AVAD & V24\_Tap4 & V36\_Guitar1 & V19\_Singing1\\  
AVAD & V13\_Conservation2 & V2\_Speech2 & V4\_Interview1\\  
AVAD & V9\_News2 & V18\_Soccer2 & V20\_Singing2\\  
AVAD & V10\_News3 & V39\_Guitar4 & V12\_Conservation1\\  
Coutrot 1 & clip4 &  clip1 & clip2\\  
Coutrot 1 & clip5 & clip3 & clip6\\  
Coutrot 1 & clip10 & clip7 & clip9\\  
Coutrot 1 & clip14 & clip8 & clip12\\  
Coutrot 1 & clip15 & clip11 & clip13\\  
Coutrot 1 & clip16 & clip25 & clip18\\  
Coutrot 1 & clip17 & clip27 & clip19\\  
Coutrot 1 & clip22 & clip28 & clip20\\  
Coutrot 1 & clip24 & clip29 & clip21\\  
Coutrot 1 & clip26 & clip30 & clip23\\  
Coutrot 1 & clip32 & clip31 & clip34\\  
Coutrot 1 & clip37 & clip33 & clip35\\  
Coutrot 1 & clip38 & clip36 & clip40\\  
Coutrot 1 & clip39 & clip42 & clip41\\  
Coutrot 1 & clip44 & clip43 & clip45\\  
Coutrot 1 & clip47 & clip49 & clip46\\  
Coutrot 1 & clip48 & clip52 & clip51\\  
Coutrot 1 & clip50 & clip54 & clip53\\  
Coutrot 1 & clip56 & clip55 & clip59\\  
Coutrot 1 & clip58 & clip57 & clip60\\  
Coutrot2 & clip4 & clip12 & clip3\\  
Coutrot2 & clip15 & clip13 & clip1\\  
Coutrot2 & clip11 & clip10 & clip9\\  
Coutrot2 & clip14 & clip7 & clip6\\  
Coutrot2 & clip2 & clip5 & clip8\\  
DIEM & BBC\_life\_in\_cold\_blood & BBC\_life\_in\_cold\_blood & BBC\_life\_in\_cold\_blood \\  
DIEM & BBC\_wildlife\_serpent & BBC\_wildlife\_serpent & BBC\_wildlife\_serpent \\  
DIEM & DIY\_SOS & DIY\_SOS & DIY\_SOS \\  
DIEM & advert\_bbc4\_bees & advert\_bbc4\_bees & advert\_bbc4\_bees \\  
DIEM & advert\_bbc4\_library & advert\_bbc4\_library & advert\_bbc4\_library \\  
DIEM & advert\_iphone & advert\_iphone & advert\_iphone \\  
DIEM & ami\_ib4010\_closeup & ami\_ib4010\_closeup & ami\_ib4010\_closeup \\  
DIEM & ami\_ib4010\_left & ami\_ib4010\_left & ami\_ib4010\_left \\  
DIEM & harry\_potter\_6\_trailer & harry\_potter\_6\_trailer & harry\_potter\_6\_trailer \\  
DIEM & music\_gummybear & music\_gummybear & music\_gummybear \\  
DIEM & music\_trailer\_nine\_inch\_nails & music\_trailer\_nine\_inch\_nails & music\_trailer\_nine\_inch\_nails \\  
DIEM & news\_tony\_blair\_resignation & news\_tony\_blair\_resignation & news\_tony\_blair\_resignation \\  
DIEM & nightlife\_in\_mozambique & nightlife\_in\_mozambique & nightlife\_in\_mozambique\\  
 \end{tabular}
}
\end{table*}

\begin{table*}[t!]
 \centering{
 \begin{tabular}{ | c | c | c | c | }
DIEM & one\_show & one\_show & one\_show \\  
DIEM & pingpong\_angle\_shot & pingpong\_angle\_shot & pingpong\_angle\_shot \\  
DIEM & pingpong\_no\_bodies & pingpong\_no\_bodies & pingpong\_no\_bodies \\  
DIEM & sport\_scramblers & sport\_scramblers & sport\_scramblers \\  
DIEM & sport\_wimbledon\_federer\_final & sport\_wimbledon\_federer\_final & sport\_wimbledon\_federer\_final \\  
DIEM & tv\_uni\_challenge\_final & tv\_uni\_challenge\_final & tv\_uni\_challenge\_final \\  
DIEM & university\_forum\_construction\_ionic & university\_forum\_construction\_ionic & university\_forum\_construction\_ionic \\  
SumMe & Valparaiso\_Downhill & Cooking & Base\_jumping \\  
SumMe & Car\_railcrossing & Bus\_in\_Rock\_Tunnel & Bike\_Polo\\  
SumMe & Bearpark\_climbing & Uncut\_Evening\_Flight & Scuba\\  
SumMe & Playing\_on\_water\_slide & playing\_ball & paluma\_jump\\  
SumMe & Fire\_Domino & St\_Maarten\_Landing & Kids\_playing\_in\_leaves\\  
SumMe & Cockpit\_Landing & Paintball & Eiffel\_Tower\\  
SumMe & Saving\_dolphines & Air\_Force\_One & Statue\_of\_Liberty\\  
SumMe & Notre\_Dame & car\_over\_camera & Excavators\_river\_crossing\\  
SumMe & & & Jumps\\  
ETMD & CHI\_1\_color & CRA\_1\_color & DEP\_1\_color\\  
ETMD & CHI\_2\_color & CRA\_2\_color & DEP\_2\_color\\  
ETMD & GLA\_1\_color & FNE\_1\_color & LOR\_1\_color\\  
ETMD & GLA\_2\_color & FNE\_2\_color & LOR\_2\_color\\  \hline
\end{tabular}
\caption{Detailed list of the video contents of each one of the three test splits. Note that regarding DIEM, the same videos are contained in every split, because there is a specific train, validation and test split from the literature.}}
\label{table:splits}
\end{table*}

\subsection{Ablation study per database}

Due to space restrictions, in the main body of the paper paper, the ablation study was presented as a table summarizing the results for all the videos contained in the employed databases. 
Here, in Tables~\ref{table:sal_eval},\ref{table:sal_eval2} the ablation study results and the state-of-the-art evaluation results have been concatenated and presented per database for completeness. For details about the state-of-the-art methods please refer to the main paper.

We notice that except for a few cases, audiovisual combinations outperform all other visual-only methods, including our visual only variant. Sometimes, the performance is better by a large margin, as for example in Coutrot2, which is the most ``audiovisual" database. In Coutrot2, all audiovisual combinations outperform the visual-only by far, indicating that the network indeed learns to fuse auditory saliency in order to predict fixations closer to the human ones.

An interesting remark concerns SumMe database, which contains the most unedited, user-made videos. Many of its videos include footages, GoPro cameras, videos with artificial sound, etc., thus, the majority does not contain many actual audiovisual events. However, audiovisual methods have performed quite well. Surprisingly though, the best performance in 3 out of 5 metrics is achieved by an audio-only method, that localizes the sound source among the other redundant information included in SumMe videos.
Regarding the rest 2 metrics, the best performance is achieved by two different state-of-the-art spatial-only visual saliency methods.

\subsection{Video demos of the proposed STAViS Network}

In the supplementary material folder we have also included 4 demo videos (which correspond to the figures of the main body of the paper) that demonstrate better our approach over several video from all the employed databases.

\begin{table*}[h]
 \centering{
 \resizebox{0.98\textwidth}{!}{
 \begin{tabular}{ l | c c c c c || c c c c c || c c c c c}
 \hline
 \multirow{2}{*}{\backslashbox{ \kern-0.5em Method \kern-0.5em}{\kern-1.9em Dataset \kern-0.5em}} &  \multicolumn{5}{c||}{DIEM}   &  \multicolumn{5}{c||}{Coutrot1} & \multicolumn{5}{c}{Coutrot2} \\ \cline{2-16} 
 & CC $\uparrow$ & NSS $\uparrow$ & AUC-J $\uparrow$ & sAUC $\uparrow$ & SIM  $\uparrow$ & CC $\uparrow$ & NSS $\uparrow$ & AUC-J $\uparrow$ & sAUC $\uparrow$ & SIM  $\uparrow$ & CC $\uparrow$ & NSS $\uparrow$ & AUC-J $\uparrow$ & sAUC $\uparrow$ & SIM  $\uparrow$\\ \hline \hline
Visual & 0.5665 & 2.19 & 0.8792 & 0.6648 & 0.4719 & 
0.4587 & 1.99 & 0.8617 & 0.5764 & 0.3842 &
0.6529 & 4.19 & 0.9405 & 0.6895 & 0.4470  \\ \hline
$L_1$\_AudioOnly & 0.5362 & 2.05 & 0.8719	& 0.6596 & 0.4573 & 
0.4444 & 1.93 & 0.8605 & 0.5789 & 0.3813 & 
0.6917 & 4.67 & 0.9519 & 0.7013 & 0.4970 \\ 
$L_2$\_AudioOnly & 0.5458 & 2.10 & 0.8737 & 0.6601 & 0.4569 & 
0.4687 & 2.04 & 0.8669 & 0.584 & 0.3880 & 
0.7223 & 4.99 & 0.9572 & 0.7054 & 0.5000\\ 
$L_3$\_AudioOnly & 0.5407 & 2.10 & 0.8719 & 0.6594 & 0.4552 &
0.4491 & 1.99 & 0.8618 & 0.5799 & 0.3745 & 
0.7126 & 4.96 & 0.9568 & 0.7023 & 0.4849\\ \hline
$L_1$\_$S_1^{av}$ & 0.5489 & 2.13 & 0.8743 & 0.6582 & 0.4623 & 
0.4516 & 2.01 & 0.8612 & 0.5803 & 0.3808 & 
0.7102 & 5.01 & 0.9523 & 0.7023 & 0.4911\\ 
$L_2$\_$S_1^{av}$ & 0.5731 & 2.25 & 0.8839 & 0.6701 & 0.4847 &
0.4577 & 2.04 & 0.8602 & 0.5770 & 0.3892 & 
0.7040 & 5.02 & 0.9480 & 0.6954 & 0.5017\\ 
$L_3$\_$S_1^{av}$ & 0.5712 & 2.25 & 0.8825 & 0.6693 & 0.4848 & 
0.4623 & 2.07 & 0.8649 & 0.5820 & 0.3934 &
0.7076 & 5.08 & 0.9499 & 0.7003 & 0.5112\\ \hline
$L_1$\_$S_2^{av}$ & 0.525 & 2.04 & 0.8367 & 0.661 & 0.3519 & 
0.4437 & 1.91 & 0.8354 & 0.5787 & 0.2994 & 
0.6250 & 3.90 & 0.9168 & 0.6958 & 0.2785\\ 
$L_2$\_$S_2^{av}$ & 0.5259 & 2.04 &	0.8376 & 0.6607 & 0.3527 & 
0.4448 & 1.92 & 0.8374 & 0.5789 & 0.3005 &
0.6257 & 3.90 & 0.9171 & 0.6952 & 0.2801\\ 
$L_3$\_$S_2^{av}$ & 0.5309 & 2.07 & 0.8344 & 0.6648 & 0.3553 & 
0.4364 & 1.90 & 0.8218 & 0.5782 & 0.2924 &
0.6309 & 4.07 & 0.9161 & 0.6977 & 0.2883\\  
$L_3^{mul}$\_$\tilde{S}_2^{av}$ &  0.5594 & 2.14 & 0.8785 & 0.6681 & 0.4694 & 0.4588 & 2.00 & 0.8633 & 0.5823 & 0.3872 &
0.6983 & 4.70 & 0.9513 & 0.7035 & 0.4762\\ \hline
$\mathbf{L_3^{mul}}$\_$\mathbf{S_3^{av}}$ (proposed) & \textbf{0.5795} & \textbf{2.26} & 0.8838 & \textbf{0.6741} & 0.4824 &
0.4722 & 2.11 & \textbf{0.8686} & \textbf{0.5847} & 0.3935 & 
\textbf{0.7349} & \textbf{5.28} & \textbf{0.9581} & \textbf{0.7106} & 0.5111\\ \hline
$L_3^{mul}$\_$S_{fus}^{av}$ & \textbf{0.5795} & 2.25 & \textbf{0.8843} & 0.6727 & \textbf{0.4877} &
0.4679 & 2.08 & 0.8670 & 0.5835 & \textbf{0.3954} &
0.7215 & 5.10 & 0.9551 & 0.7083 & \textbf{0.5137}\\ \hline \hline
DeepNet~\cite{Pan_2016_CVPR}& 0.4075 & 1.52 & 0.8321 & 0.6227 & 0.3183& 
0.3402 & 1.41 & 0.8248 & 0.5597 & 0.2732 & 
0.3012 & 1.82 & 0.8966 & 0.6000 & 0.2019 \\
DVA~\cite{wang2018deep} & 0.4779 & 1.97 & 0.8547 & 0.641 & 0.3785 & 
0.4306 & 2.07 & 0.8531 & 0.5783 & 0.3324
& 0.4634 & 3.45 & 0.9328 & 0.6324 & 0.2742 \\
SAM~\cite{cornia2018predicting}  & 0.4930 & 2.05 & 0.8592 & 0.6446 & 0.4261 &
0.4329 & 2.11 & 0.8571 & 0.5768 & 0.3672 & 
0.4194 & 3.02 & 0.9320 & 0.6152 & 0.3041 \\ 
SalGAN~\cite{pan2017salgan} & 0.4868 & 1.89 & 0.8570 & 0.6609 & 0.3931 &
0.4161 & 1.85 & 0.8536 & 0.5799 & 0.3321 & 
0.4398 & 2.96 & 0.9331 & 0.6183 & 0.2909 \\
ACLNet~\cite{wang2018revisiting,wang2019revisiting} & 0.5229 & 2.02 & 0.8690 & 0.6221 & 0.4279 & 
0.4253 & 1.92 & 0.8502 & 0.5429 & 0.3612 & 
0.4485 & 3.16 & 0.9267 & 0.5943 & 0.3229 \\
DeepVS~\cite{jiang2018deepvs} & 0.4523 & 1.86 & 0.8406 & 0.6256 & 0.3923 &
0.3595 & 1.77 & 0.8306 & 0.5617 & 0.3174 & 
0.4494 & 3.79 & 0.9255 & 0.6469 & 0.2590 \\ 
TASED~\cite{Min_2019_ICCV} & 0.5579 & 2.16 & 0.8812 & 0.6579 & 0.4615 &
\textbf{0.4799} & \textbf{2.18} & 0.8676 & 0.5808 & 0.3884 & 
0.4375 & 3.17 & 0.9216 & 0.6118 & 0.3142 \\ \hline 
\end{tabular}
}}
\vspace{0.3cm}
\caption{Ablation study and state-of-the-art evaluation for databases DIEM, Coutrot1 and Coutrot2.}
\label{table:sal_eval}
\end{table*}

\begin{table*}[h]
 \centering{
 \resizebox{0.98\textwidth}{!}{
 \begin{tabular}{ l | c c c c c || c c c c c || c c c c c}
 \hline
 \multirow{2}{*}{\backslashbox{ \kern-0.5em Method \kern-0.5em}{\kern-1.9em Dataset \kern-0.5em}} &  \multicolumn{5}{c||}{AVAD}   &  \multicolumn{5}{c||}{SumMe} & \multicolumn{5}{c}{ETMD} \\ \cline{2-16} 
 & CC $\uparrow$ & NSS $\uparrow$ & AUC-J $\uparrow$ & sAUC $\uparrow$ & SIM  $\uparrow$ & CC $\uparrow$ & NSS $\uparrow$ & AUC-J $\uparrow$ & sAUC $\uparrow$ & SIM  $\uparrow$ & CC $\uparrow$ & NSS $\uparrow$ & AUC-J $\uparrow$ & sAUC $\uparrow$ & SIM  $\uparrow$\\ \hline \hline
Visual & 0.6041 & 3.07 & 0.9157 & 0.5900 & 0.4431 & 
0.4180 & 1.98 & 0.8848 & 0.6477 & 0.3325 &
0.5602 & 2.84 & 0.9290 & 0.7278 & 0.4121  \\ \hline
$L_1$\_AudioOnly & 0.5836 & 2.89 & 0.9176 & 0.5949 & 0.4413 &
0.4072 & 1.93 & 0.8833 & 0.6520 & 0.3328 & 
0.5407 & 2.71 & 0.9276 & 0.7298 & 0.4086\\ 
$L_2$\_AudioOnly & 0.6107 & 3.07 & 0.9199 & 0.5975 & 0.4475 & 
\textbf{0.4413} & 2.12 & \textbf{0.8908} & 0.6665 & \textbf{0.3442} &
0.5580 & 2.84 & 0.9302 & 0.7343 & 0.4087\\ 
$L_3$\_AudioOnly & 0.6009 & 3.07& 0.9185 & 0.5956 & 0.4351 &
0.4097 & 1.95 & 0.8838 & 0.6527 & 0.3218 &
0.5504 & 2.81 & 0.9285 & 0.7301 & 0.3995\\ \hline
$L_1$\_$S_1^{av}$ & 0.6035 & 3.13& 0.9166 & \textbf{0.5984} & 0.4463 & 
0.4069 & 1.94 & 0.8826 & 0.6463 & 0.3271 &
0.5493 & 2.82 & 0.9277 & 0.7264 & 0.4073\\ 
$L_2$\_$S_1^{av}$ & \textbf{0.6198} & \textbf{3.26} & 0.9201 & 0.5949 & \textbf{0.4700} &
0.4146 & 2.01 & 0.8872 & 0.6514 & 0.3359 &
0.5599 & 2.90 & 0.9303 & 0.7256 & 0.4238\\ 
$L_3$\_$S_1^{av}$ & 0.6159 & 3.25 & 0.9195 & 0.5934 & 0.4690 &
0.4136 & 1.99 & 0.8856 & 0.6520 & 0.3391 &
0.5658 & \textbf{2.93} & 0.9312 & 0.7292 & 0.4296\\ \hline
$L_1$\_$S_2^{av}$ & 0.5886 & 2.96 & 0.9013 & 0.5951 & 0.3156 &
0.4035 & 1.91 & 0.8618 & 0.6540 & 0.2470 &
0.5418 & 2.74 & 0.911 & 0.7346 & 0.2773\\ 
$L_2$\_$S_2^{av}$ & 0.5912 & 2.98 & 0.9021 & 0.5957 & 0.3172 & 
0.4038 & 1.91 & 0.8627 & 0.6542 & 0.2476 &
0.5423 & 2.74 & 0.9116 & 0.7346 & 0.2784\\ 
$L_3$\_$S_2^{av}$ & 0.5860 & 2.96 & 0.8899 & 0.5947 & 0.3036 &
0.4110 & 1.97 & 0.8710 & 0.6513 & 0.2620 &
0.5530 & 2.86 & 0.9161 & 0.7311 & 0.3070\\  
$L_3^{mul}$\_$\tilde{S}_2^{av}$ & 0.5966 & 2.98 & 0.9179 & 0.5914 & 0.4449 &
0.4188 & 1.99 & 0.8853 & 0.6566 & 0.3358 &
0.5551 & 2.80 & 0.9292 & 0.7333 & 0.4136\\ \hline
$\mathbf{L_3^{mul}}$\_$\mathbf{S_3^{av}}$ (proposed)& 0.6086 & 3.18 & 0.9196 & 0.5936 & 0.4578 &
0.4220 & 2.03 & 0.8883 & 0.6562 & 0.3373 &
\textbf{0.5690} & \textbf{2.93} & 0.9316 & 0.7317 & 0.4251\\ \hline
$L_3^{mul}$\_$S_{fus}^{av}$ & 0.6162 & 3.21 & \textbf{0.9216} & 0.5952 & 0.4690 &
0.4183 & 2.01 & 0.8871 & 0.6547 & 0.3403 &
0.5669 & 2.91 & \textbf{0.9317} & 0.7318 & \textbf{0.4280}\\ \hline \hline
DeepNet~\cite{Pan_2016_CVPR} & 0.3831 & 1.85 & 0.8690 & 0.5616 & 0.2564& 
0.3320 & 1.55 & 0.8488 & 0.6451 & 0.2274 & 
0.3879 & 1.90 & 0.8897 & 0.6992 & 0.2253 \\
DVA~\cite{wang2018deep} & 0.5247 & 3.00 & 0.8887 & 0.5820 & 0.3633 & 
0.3983 & 2.14 & 0.8681 & 0.6686 & 0.2811
& 0.4965 & 2.72 & 0.9039 & 0.7288 & 0.3165 \\
SAM~\cite{cornia2018predicting} & 0.5279 & 2.99 & 0.9025 & 0.5777 & 0.4244 &
0.4041 & \textbf{2.21} & 0.8717 & 0.6728 & 0.3272 & 
0.5068 & 2.78 & 0.9073 & 0.7310 & 0.3790 \\ 
SalGAN~\cite{pan2017salgan}  & 0.4912 & 2.55 & 0.8865 & 0.5799 & 0.3608 &
0.3978 & 1.97 & 0.8754 & \textbf{0.6882} & 0.2897 & 
0.4765 & 2.46 & 0.9035 & \textbf{0.7463} & 0.3117 \\ 
ACLNet~\cite{wang2018revisiting,wang2019revisiting} & 0.5809 & 3.17 & 0.9053 & 0.5600 & 0.4463 & 
0.3795 & 1.79 & 0.8687 & 0.6092 & 0.2965 & 
0.4771 & 2.36 & 0.9152 & 0.6752 & 0.3290 \\
DeepVS \cite{jiang2018deepvs} & 0.5281 & 3.01 & 0.8968 & 0.5858 & 0.3914 &
0.3172 & 1.62 & 0.8422 & 0.6120 & 0.2622 & 
0.4616 & 2.48 & 0.9041 & 0.6861 & 0.3495 \\ 
TASED~\cite{Min_2019_ICCV} & 0.6006 & 3.16 & 0.9146 & 0.5898 & 0.4395 &
0.4288 & 2.10 & 0.8840 & 0.6570 & 0.3337 & 
0.5093 & 2.63 & 0.9164 & 0.7117 & 0.3660 \\ \hline 
\end{tabular}
}}
\vspace{0.3cm}
\caption{Ablation study and state-of-the-art evaluation for databases AVAD, SumMe and ETMD.}
\label{table:sal_eval2}
\end{table*}

\end{document}